\documentclass{article}

\PassOptionsToPackage{numbers,compress}{natbib}
\usepackage[preprint]{neurips_2026}

\usepackage[utf8]{inputenc}
\usepackage[T1]{fontenc}
\usepackage{hyperref}
\usepackage{url}
\usepackage{booktabs}
\usepackage{amsfonts}
\usepackage{nicefrac}
\usepackage{microtype}
\usepackage{xcolor}

\usepackage{amsmath,amssymb}
\usepackage{algorithmic}
\usepackage{algorithm}
\usepackage{graphicx}
\usepackage{multirow}
\usepackage{enumerate}
\usepackage{listings}
\usepackage{tikz}
\usetikzlibrary{shapes,arrows,positioning,fit,calc}
\usepackage{pgfplots}
\pgfplotsset{compat=1.18}
\usepackage{tabularx}
\usepackage[capitalise]{cleveref}

\graphicspath{{./figs/}}

\newcommand{\methodname}{SEATS}
\newcommand{\fullname}{Self-Evolving Agent for Time Series Algorithms}
\newcommand{\mamcts}{MA-MCTS}

\newcommand{\eg}{\textit{e.g.}}

\title{SEATS: Self-Evolving Agent for Autonomous Code Generation of Time Series Forecasting Algorithms}

\author{  
\textbf{Longkun Xu}$^{1,\dagger}$\;
\textbf{Xiaochun Zhang}$^{1}$\;
\textbf{Qiantu Tuo}$^1$\;
\textbf{Rui Li}$^{1,\dagger}$\\
$^1$  Ecoflow Inc.\\
\textsuperscript{$\dagger$}{Corresponding Authors}\;
\\
\texttt{\{luke.xu, ricky.li\}@ecoflow.com} \\
}

\begin{document}
\maketitle
\begin{abstract}
Accurate time series forecasting underpins decision-making in many domains, yet conventional ML development often faces data scarcity, distribution shift, and diminishing returns from manual iteration. We propose \textbf{\fullname{} (\methodname{})}, a framework that autonomously generates, validates, and optimizes forecasting algorithm code through an iterative self-evolution loop. Our design combines three mechanisms: (1) \textbf{Metric-Advantage MCTS (\mamcts{})}, which replaces fixed rewards with a statistically normalized advantage score for search guidance, (2) \textbf{code review with running prompt refinement}, so every successfully executed solution is reviewed and the running prompt encodes corrective patterns for later iterations, and (3) \textbf{global steerable reasoning}, which compares each evaluated node to global best- and worst-performing solutions for cross-trajectory transfer. A MAP-Elites archive maintains architectural diversity. Across four datasets and two metrics, \methodname{} wins seven of eight comparisons against strong baselines TimeMixer~\cite{wang2024timemixer}, Timer~\cite{liu2024timer}, and SEMixer~\cite{zhang2026semixer}.
\end{abstract}

\section{Introduction}
\label{sec:introduction}

Developing high-performance machine learning models for time series forecasting remains a labor-intensive process that demands substantial domain expertise and iterative experimentation. Although the task spans numerous application domains---including energy systems~\cite{hong2016probabilistic,van2018review}, financial markets, supply chain logistics, healthcare monitoring, and environmental science---the fundamental challenges are shared across all settings. We identify three pervasive obstacles in the conventional ML development pipeline (data collection $\rightarrow$ dataset construction $\rightarrow$ model design $\rightarrow$ training $\rightarrow$ deployment):

\begin{enumerate}[(a)]
    \item \textbf{Data scarcity:} Many practical forecasting scenarios lack sufficient historical data. Newly deployed systems, emerging markets, or rare event prediction all face a prolonged cold-start period where training reliable models is infeasible~\cite{shi2017deep}. This data scarcity problem is especially severe for tasks requiring seasonal coverage (\eg, one year of data to capture annual patterns).
    
    \item \textbf{Distribution shift:} Real-world time series are inherently non-stationary. Equipment degradation, policy changes, evolving user behaviors, and environmental shifts cause the data distribution to drift continuously. Models trained on historical data degrade under these shifted conditions, necessitating frequent and costly human intervention for retraining and adaptation.
    
    \item \textbf{Diminishing marginal returns:} As models approach their performance ceiling, each incremental accuracy improvement requires disproportionately more engineering effort. For instance, once a practitioner improves accuracy from 85\% to 90\%, further gains yield diminishing returns on investment~\cite{he2021automl}, making manual iteration economically unsustainable.
\end{enumerate}

Recent advances in Large Language Models (LLMs) have opened a new paradigm: using LLMs as \textit{machine learning engineers} (MLE) that autonomously write, execute, evaluate, and iteratively improve ML code~\cite{huang2024mlagentbench, chan2024mle}. Frameworks such as AIDE~\cite{jiang2025aide} and ML-Master~\cite{mlmaster2024} have demonstrated that LLM-based agents, combined with tree search strategies, can generate competitive ML solutions across diverse tasks. However, existing approaches suffer from several limitations:

\begin{enumerate}[(a)]
    \item \textbf{Reward hacking and ``cheating code'':} Agents driven by metric optimization frequently produce code with logical flaws (\eg, data leakage in time series tasks) that yield artificially inflated scores but fail in deployment. Such flaws do not cause runtime errors, making them difficult to detect without explicit code review~\cite{browne2012survey}. Such issue is also reported in recent works of MLE agent~\cite{singh2026agentic}.
    
    \item \textbf{Simplistic reward mechanisms:} Existing approaches typically use binary or fixed rewards (\eg, $+1$ for any improvement~\cite{mlmaster2024}), which fail to distinguish marginal gains from significant breakthroughs, leading to inefficient search.
    
    \item \textbf{Limited reasoning context:} Most agents reference only local information (parent/sibling nodes) when generating new code, lacking global awareness of best and worst solutions across the entire search history.
    
    \item \textbf{Static prompts:} System prompts remain fixed throughout the search, missing opportunities for on-the-fly adaptation based on discovered failure modes or successful patterns.
\end{enumerate}

To address these challenges, we propose the \textbf{\fullname{} (\methodname{})} framework with three synergistic innovations:

\begin{enumerate}[(a)]
    \item \textbf{Metric-Advantage Monte Carlo Tree Search (\mamcts{}):} We replace fixed rewards with a statistically normalized advantage score computed from the full metric distribution, enabling discriminative guidance toward high-potential trajectories.
    
    \item \textbf{Code Review with running prompt refinement:} Every successfully executed solution undergoes automated logical review. The review findings---together with insights from global node comparisons---are distilled into a continuously evolving \textit{running prompt} that prevents recurrence of identified issues in all subsequent iterations.
    
    \item \textbf{Global Steerable Reasoning:} We compare each evaluated node against the global best and worst solutions, enabling cross-trajectory knowledge transfer and avoiding redundant exploration.
\end{enumerate}

Additionally, we integrate a MAP-Elites quality-diversity archive~\cite{mouret2015illuminating} to maintain diverse elite solutions across architectural dimensions. As also reported in other work, it is critical to enforce diversity via prompt tuning\cite{wang2026self}. We evaluate \methodname{} on energy forecasting as a representative time series domain and report comparisons against strong published baselines (\Cref{sec:results}). SEATS can design nontrivial architectural patterns---\eg, physics-motivated decay components and per-station diurnal structure---that are not provided in the initial prompt.

The main contributions of this paper are:
\begin{enumerate}[(a)]
    \item A general-purpose self-evolving MLE agent framework unifying metric-advantage tree search, automated code review with running prompt refinement, and global steerable reasoning for autonomous algorithm development.
    \item Empirical comparison on four energy forecasting settings shows that self-evolved code is preferred to strong published baselines on seven out of eight comparisons, together with ablations verified on the public ECL dataset.
\end{enumerate}

The remainder of the paper is organized as follows. \Cref{sec:related_work} reviews related work. \Cref{sec:methodology} formalizes the problem and presents the \methodname{} framework. \Cref{sec:experiments} describes datasets, baselines, and empirical results. \Cref{sec:conclusion} concludes and discusses future work. Pseudocode, token and wall-clock analysis, supplementary tables, and extended discussion appear in the appendix.

\section{Related Work}
\label{sec:related_work}

\subsection{Automated Machine Learning}

Traditional AutoML approaches, including neural architecture search (NAS)~\cite{zoph2017neural,liu2018darts} and hyperparameter optimization~\cite{feurer2015efficient}, operate over predefined architectural spaces. LLM-based agents instead generate executable programs with fewer structural constraints---including novel feature engineering, custom losses, and domain inductive biases---at a different abstraction level than fixed search spaces.

\subsection{LLM-Based Machine Learning Engineering Agents}

Powerful LLMs enable systems where the model acts as a programmer. AIDE~\cite{jiang2025aide} pioneered greedy search over code, ML-Master~\cite{mlmaster2024} incorporated MCTS and steerable reasoning~\cite{zhu2026toward}, I-MCTS~\cite{liang2025mcts} introduced hybrid rewards, R\&D-Agent~\cite{yang2025r} combined multi-trajectory exploration with a multi-agent framework. Common limitations include vulnerability to reward hacking without explicit domain-aware review, limited reward granularity, and reasoning context that stays local to a branch. \methodname{} is designed to address these gaps jointly.

\subsection{Quality-Diversity Optimization}

MAP-Elites~\cite{mouret2015illuminating} and variants maintain archives of diverse high-performing solutions indexed by behavioral features. AlphaEvolve~\cite{novikov2025alphaevolve} and OpenEvolve apply quality-diversity ideas to LLM program synthesis. We adapt MAP-Elites to preserve architectural diversity among generated ML pipelines.

\subsection{Time Series Forecasting Methods}

Forecasting has been studied from classical statistics to deep learning. Transformers and related architectures are widely used for long horizons~\cite{zhou2021informer,wu2021autoformer,nie2022time}, with ongoing debate on inductive biases~\cite{zeng2023transformers}. TimeMixer~\cite{wang2024timemixer} emphasizes multi-scale mixing, iTransformer~\cite{liu2024itransformer} inverts the attention pattern. Domain applications often combine physics with data-driven components~\cite{das2018forecasting,lai2017daily,hong2016probabilistic,wang2018review}. Adapting such models to new deployments nonetheless remains labor-intensive, motivating automated algorithm search.

\section{Methodology}
\label{sec:methodology}

\subsection{Problem Formulation}
\label{sec:problem_formulation}

We formalize autonomous algorithm development as a search problem over the space of programs. Let $\Pi$ denote the set of all syntactically valid ML programs that can be generated by an LLM. Given a dataset $\mathcal{D} = (\mathcal{D}_{\text{train}}, \mathcal{D}_{\text{val}}, \mathcal{D}_{\text{test}})$ and an evaluation metric $\mathcal{L}: \Pi \times \mathcal{D} \to \mathbb{R}$, the objective is:
\begin{equation}
\pi^* = \arg\min_{\pi \in \Pi} \,\mathcal{L}\!\left(\pi(\mathcal{D}_{\text{train}}),\, \mathcal{D}_{\text{test}}\right),
\label{eq:objective}
\end{equation}
\noindent where $\pi(\mathcal{D}_{\text{train}})$ denotes the trained model produced by executing program $\pi$ on training data, and $\mathcal{L}$ is task-dependent (\eg, WAPE, MAPE, or MAE).

Since $\Pi$ is combinatorially vast and lacks tractable gradients, we model the search as a sequential decision process over a tree $\mathcal{T} = (\mathcal{V}, \mathcal{E})$, where each vertex $N_j \in \mathcal{V}$ represents a complete solution state and each edge $(N_i, N_j) \in \mathcal{E}$ represents a refinement operation. The root $N_0$ contains an initial reference implementation or virtue node when reference code is not available.

\subsection{Framework Overview and Self-Evolution Loop}
\label{sec:overview}

\begin{figure}[htbp]
    \centering
    \includegraphics[width=1.0\linewidth]{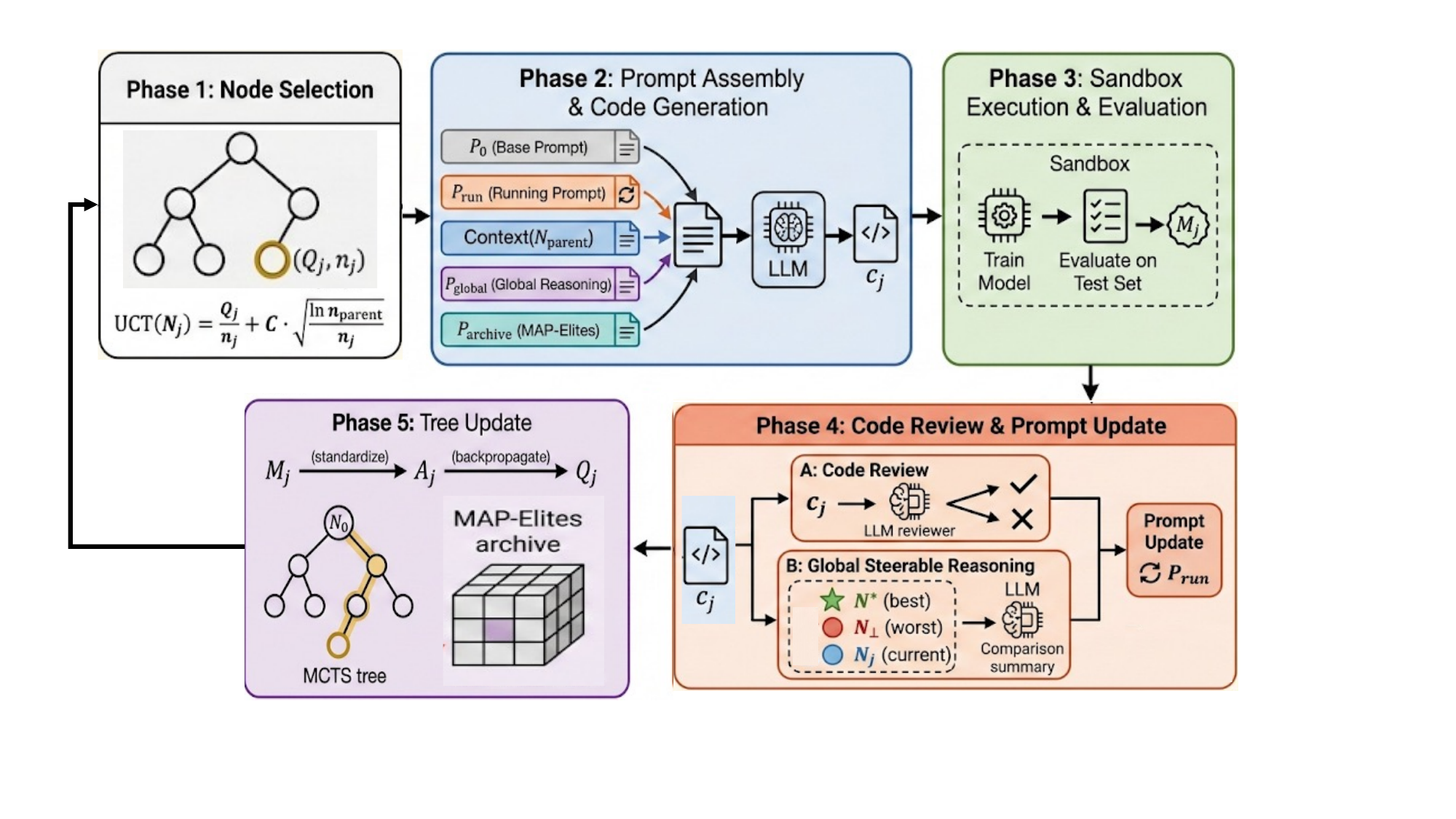}    
    \caption{Overview of the \methodname{} framework. The system iteratively generates, evaluates, and refines ML code through a closed-loop combining MA-MCTS search, code review with running prompt refinement, and global steerable reasoning.}
    \label{fig:framework}
\end{figure}

The \methodname{} framework operates as a closed-loop self-evolution system (\Cref{fig:framework}). At each iteration $t \in \{1, \ldots, T\}$, the system executes five phases: (1) \textit{Node Selection} via UCT, (2) \textit{Prompt Assembly \& Code Generation}, (3) \textit{Sandbox Execution \& Evaluation}, (4) \textit{Code Review \& Prompt Update} for every non-buggy node, and (5) \textit{Tree Update} including reward computation, backpropagation, and archive maintenance.

The prompt for code generation is composed as:
\begin{equation}
\mathcal{P}(N_{\text{parent}}, \mathcal{T}) = \mathcal{P}_0 \oplus \mathcal{P}_{\text{run}} \oplus \textsc{Context}(N_{\text{parent}}) \oplus \mathcal{P}_{\text{global}}(\mathcal{T}) \oplus \mathcal{P}_{\text{archive}}(\mathcal{A}),
\label{eq:prompt}
\end{equation}
\noindent where $\mathcal{P}_0$ is the base task description (fixed), $\mathcal{P}_{\text{run}}$ is the accumulated running prompt (continuously updated), $\textsc{Context}(N_{\text{parent}})$ provides parent and sibling node information, $\mathcal{P}_{\text{global}}$ contains global best/worst comparisons, and $\mathcal{P}_{\text{archive}}$ samples from the MAP-Elites archive $\mathcal{A}$. Illustrative prompt templates are provided in \Cref{app:prompts}.

Figure~\ref{fig:framework} summarizes the loop, step-by-step pseudocode is deferred to \Cref{app:algorithm}.

\subsection{Metric-Advantage Monte Carlo Tree Search (MA-MCTS)}
\label{sec:mamcts}

\subsubsection{Search Tree Structure}

Each node $N_j$ in the search tree represents a \textit{complete solution state}. While the code $c_j$ is the executable artifact---a self-contained Python program---the node $N_j$ encapsulates additional metadata:
\begin{enumerate}
    \item Code $c_j \in \Sigma^*$: the full Python ML program (immutable once created)
    \item Plan $p_j$: the design rationale and approach description
    \item Metric $M_j \in \mathbb{R}$: the evaluation performance (\eg, MAE, MSE)
    \item Visit count $n_j \in \mathbb{N}$ and cumulative value $Q_j \in \mathbb{R}$, see Figure ~\ref{fig:framework}
    \item Buggy flag $b_j \in \{\text{true}, \text{false}\}$: whether code was identified as logically flawed
\end{enumerate}

The search algorithm operates on nodes (selecting, expanding, backpropagating), while the LLM and sandbox operate on code (generating, executing, evaluating). A node's code $c_j$ is immutable, but its MCTS statistics ($n_j$, $Q_j$) are continuously updated via backpropagation.

\subsubsection{Node Selection via UCT}

The agent selects a leaf node for expansion using the standard Upper Confidence Bound for Trees (UCT):
\begin{equation}
\text{UCT}(N_j) = \underbrace{\frac{Q_j}{n_j}}_{\text{exploitation}} + \underbrace{C \sqrt{\frac{\ln n_{\text{parent}}}{n_j}}}_{\text{exploration}},
\label{eq:uct}
\end{equation}
\noindent where $C > 0$ is the exploration constant. The default option in this work is using a fixed constant $C$ (1.41) while it can be decayed over the budget to shift from exploration to exploitation.

\subsubsection{Metric Advantage Reward and the $M$--$A$--$R$--$Q$ Chain}
\label{sec:reward_chain}

A key innovation is the \textbf{Metric Advantage Reward}, which replaces fixed rewards with a statistically normalized signal. The four key quantities form a causal chain:
\begin{equation}
M_j \xrightarrow{\text{standardize}} A_j \xrightarrow{\text{review}} R_j \xrightarrow{\text{backprop}} Q_j
\label{eq:chain}
\end{equation}

\paragraph{Step 1: Metric $\boldsymbol{M_j}$.} The raw evaluation metric obtained by executing code $c_j$:
\begin{equation}
M_j = \mathcal{L}(c_j(\mathcal{D}_{\text{train}}), \mathcal{D}_{\text{test}}).
\label{eq:metric}
\end{equation}

\paragraph{Step 2: Advantage $\boldsymbol{A_j}$.} Given historical metrics $\mathcal{M} = \{M_1, \ldots, M_n\}$, the advantage is the standardized z-score:
\begin{equation}
A_j = \begin{cases}
\displaystyle\frac{\mu(\mathcal{M}) - M_j}{\sigma(\mathcal{M})} & \text{if lower is better (\eg, MAE)} \\[8pt]
\displaystyle\frac{M_j - \mu(\mathcal{M})}{\sigma(\mathcal{M})} & \text{if higher is better (\eg, accuracy)}
\end{cases}
\label{eq:advantage}
\end{equation}
\noindent where $\mu(\mathcal{M})$ and $\sigma(\mathcal{M})$ are the mean and standard deviation of the historical multiset $\mathcal{M}$ (requiring $\sigma(\mathcal{M})>0$). When $\mu$ and $\sigma$ are computed from that same multiset, the standardized values have zero mean and unit variance over $\mathcal{M}$, yielding discriminative ($A_j \gg 1$ for breakthroughs), adaptive (scale adjusts as $\mathcal{M}$ evolves), and metric-agnostic reward signals.

\paragraph{Step 3: Reward $\boldsymbol{R_j}$.} The final reward incorporates the code review outcome:
\begin{equation}
R_j = \begin{cases}
-1 & \text{if } b_j = \text{true} \\
A_j & \text{otherwise}
\end{cases}
\label{eq:reward}
\end{equation}
\noindent where $b_j$ can be set during execution (runtime errors) or by the code review (\Cref{sec:code_review}).

\paragraph{Step 4: Cumulative Value $\boldsymbol{Q_j}$.} When node $N_k$ is created as a descendant, its reward $R_k$ is backpropagated:
\begin{equation}
Q_j \leftarrow Q_j + R_k, \quad n_j \leftarrow n_j + 1, \quad \forall\, N_j \in \text{path}(N_k, N_0).
\label{eq:backprop}
\end{equation}
\noindent Thus $Q_j = \sum_{k \in \text{desc}(j)} R_k$, and UCT uses $Q_j / n_j$ as the exploitation term. Breakthrough solutions ($A_j \gg 1$) boost $Q_j/n_j$ for all ancestors, buggy code ($R_j = -1$) penalizes the entire ancestor path. As $\mathcal{M}$ evolves and $\sigma(\mathcal{M})$ shrinks, the same absolute improvement yields larger $A_j$, naturally intensifying exploitation as the search matures.

\subsection{Code Review with Running Prompt Refinement}
\label{sec:code_review}

\subsubsection{Motivation}

In time series forecasting, a subtle form of reward hacking is \textit{data leakage}---using future information in features (\eg, rolling statistics without a proper time shift). Such bugs need not crash execution but can inflate metrics. We therefore apply a dedicated logical review with domain-appropriate checks.

\subsubsection{Automated Code Review}

After each candidate $N_j$ executes without runtime errors, an LLM reviewer analyzes \textit{every} non-buggy node. The checklist includes leakage in feature construction, incorrect inverse normalization at evaluation time, train--test contamination, inference bugs, and checkpoint mismatches. The reviewer outputs $b_j \in \{\text{true}, \text{false}\}$ (true indicates a detected flaw). We err on the side of rejecting suspicious code so that flawed solutions do not inflate $Q_j$ through backpropagation.

\subsubsection{Running Prompt Refinement}

The \textit{running prompt} $\mathcal{P}_{\text{run}}$ is updated after each review and after global comparisons (\Cref{sec:global_reasoning}): an auxiliary LLM turns findings into concise, actionable rules. For example, after detecting leakage, the prompt may record that rolling features must use \texttt{.shift(1)} before windowed aggregation. Over time, $\mathcal{P}_{\text{run}}$ accumulates both failure modes and successful patterns shared across branches.

\subsection{Global Steerable Reasoning}
\label{sec:global_reasoning}

\subsubsection{Beyond Local Context}

Standard MCTS-based coding agents often condition on parent and sibling nodes only~\cite{mlmaster2024}. We augment this with a \textit{global} signal: each newly evaluated non-buggy node is compared to the current global best $N^*$ and worst $N_\bot$. The resulting analysis is stored with the node so that when it becomes a parent, children benefit from that global perspective.

\subsubsection{Comparison Prompt Construction}

We summarize the trio $(c^*, p^*, M^*)$, $(c_{\bot}, p_{\bot}, M_{\bot})$, and $(c_j, p_j, M_j)$ in a structured comparison prompt:
\begin{equation}
\mathcal{P}_{\text{global}} = \{(c^*, p^*, M^*),\, (c_{\bot}, p_{\bot}, M_{\bot}),\, (c_j, p_j, M_j)\},
\end{equation}
\noindent and ask an auxiliary LLM to contrast what works, what fails, and what to change at the code level.

\subsubsection{Cross-Trajectory Knowledge Transfer}

This mechanism enables \textit{cross-trajectory} transfer: lessons from one branch inform generation elsewhere in the tree, not only along a single refinement chain. The same summary is also distilled into $\mathcal{P}_{\text{run}}$, so global insights persist for all subsequent iterations.

\subsection{MAP-Elites Quality-Diversity Archive}
\label{sec:map_elites}

To limit collapse to a single architecture family, we maintain a MAP-Elites archive with three phenotypic dimensions:

\begin{enumerate}
    \item \textbf{Architecture type} ($d_1$): from tree-based (0.0) through decomposition (0.4), attention (0.6), to hybrid (1.0).
    \item \textbf{Feature engineering} ($d_2$): from none (0.0) to extensive (1.0), including lags, rolling statistics, and Fourier features.
    \item \textbf{Training sophistication} ($d_3$): from basic (0.0) to advanced (1.0), including schedules, regularization, and adaptive losses.
\end{enumerate}

Each grid cell retains the best solution only, yielding a compact set of diverse elites. Dimensions are configurable, learning them automatically from task metadata remains future work (\Cref{sec:conclusion}).

LLM call counts, token ranges, and wall-clock behavior are summarized in \Cref{app:complexity_details}.

\section{Experiments}
\label{sec:experiments}

\subsection{Setup}
\label{sec:setup}

\subsubsection{Datasets}

We evaluate \methodname{} on energy time series forecasting using both public benchmarks and industry proprietary datasets. While the framework is domain-agnostic, energy forecasting provides a practically important evaluation domain with well-established baselines.

\paragraph{Public Dataset.}
\textbf{Solar-Energy}~\cite{lai2018modeling}: Solar power production records from 137 PV plants in Alabama at 10-minute intervals, with standard train/val/test splits.
\textbf{ECL (Electricity Consuming Load)}: We use the standard long-horizon electricity load series from the benchmark suite associated with recent deep forecasting models~\cite{zhou2021informer,wang2024timemixer}. This dataset is also used in our ablations studies (\Cref{fig:ablation_ecl}).

\paragraph{Proprietary Datasets.}
\textbf{Proprietary PV}: Hourly solar generation from multiple distributed PV stations (Oct 2023--Mar 2025, train: Oct 2023--Oct 2024, val: Oct--Dec 2024, test: Dec 2024--Mar 2025).
\textbf{Proprietary Load}: Hourly residential electricity consumption (May--Oct 2025) with high inter-user variability.

\subsubsection{Baselines and Reference Code}

For each task, \methodname{} may start from a provided baseline reference implementation (the default is TimeMixer). We compare against published implementations of TimeMixer~\cite{wang2024timemixer}, Timer~\cite{liu2024timer}, and SEMixer~\cite{zhang2026semixer}. The reference code, when used, defines data loading, metric computation, and an initial model, the agent iteratively improves through the self-evolution loop. We additionally report \methodname{} without reference code to isolate the effect of reference code. A task--dataset summary table is in \Cref{tab:baselines}.

\subsubsection{Implementation Details}

\methodname{} uses GPT-5 (high reasoning effort) for code generation, and Qwen3-coder-plus as an alternative model for other tasks. The MCTS exploration constant is $C = \sqrt{2}$, with budget $T = 500$ iterations per task. The MAP-Elites archive uses $7^3 = 343$ cells. We include the remaining iteration count in the prompt to encourage exploration of underexplored approaches in later stages, following~\cite{gao2025more}. Experiments are conducted on Microsoft Azure cloud infrastructure. Our workflow for deep research and prompt updates across runs is described in \Cref{app:deep_research}.

\subsection{Results}
\label{sec:results}

\subsubsection{Metric Reporting}

All metrics in \Cref{tab:main_results} are on the \textit{original scale} after each method's inverse scaling, programs may use different normalizers (MAD, standard scaling, etc.). Reporting on the original scale keeps physical units interpretable~\cite{tao2026memcast}. 

\begin{table}[htbp]
\centering
\caption{Test MAE and MSE on the original scale. \textbf{Bold}: best per dataset-metric combination. ECL MSE is $\mathrm{MSE}/10^{6}$. ``Ours''\,=\,\methodname{}, ``w/ ref'' uses TimeMixer as reference.}
\label{tab:main_results}
\begingroup
\footnotesize
\setlength{\tabcolsep}{2pt}
\resizebox{\linewidth}{!}{%
\begin{tabular}{@{}l *{5}{r} *{5}{r}@{}}
\toprule
& \multicolumn{5}{c}{\textbf{MAE}} & \multicolumn{5}{c}{\textbf{MSE}} \\
\cmidrule(lr){2-6} \cmidrule(lr){7-11}
\textbf{Dataset}
  & \textbf{Ours w/ ref} & \textbf{Ours w/o ref} & \textbf{TimeMixer} & \textbf{Timer} & \textbf{SEMixer}
  & \textbf{Ours w/ ref} & \textbf{Ours w/o ref} & \textbf{TimeMixer} & \textbf{Timer} & \textbf{SEMixer} \\
\midrule
Solar-Energy (public)
  & \textbf{1.757} & 1.813 & 2.929 & 7.199 & 2.155
  & \textbf{15.653} & 16.667 & 22.682 & 118.348 & 17.496 \\
Industry solar
  & \textbf{10.408} & 29.599 & 14.259 & 37.378 & 34.980
  & \textbf{664.127} & 1549.158 & 1548.998 & 7077.054 & 6048.086 \\
ECL (public)
  & 211.625 & 209.324 & 216.489 & 245.724 & \textbf{207.993}
  & 4.926 & \textbf{4.336} & 4.793 & 5.823 & 4.999 \\
Industry load
  & \textbf{121.451} & 144.820 & 216.376 & 172.929 & 166.247
  & \textbf{153{,}957.6} & 231{,}031.0 & 317{,}860.2 & 238{,}500.7 & 219{,}572.0 \\
\bottomrule
\end{tabular}}
\endgroup
\end{table}

In \Cref{tab:main_results}, \methodname{} wins seven of eight comparisons, the remaining gap is on ECL MAE, where SEMixer reaches $207.993$ versus our $209.324$ ($\approx$0.6\%). For ECL MSE, the run without reference code achieves $4.336$, below SEMixer ($4.999$) and TimeMixer ($4.793$), with a compact archived solution ($\sim$340 lines). Although search targets MAE, MSE on the original scale also improves versus strong baselines, suggesting \textit{cross-metric generalization} rather than overfitting to a single score. In three of four datasets, ``w/ ref'' improves on both MAE and MSE versus ``w/o ref'', indicating that a strong reference seed can materially help self-evolution.

\begin{figure}[htbp]
\centering
\includegraphics[width=\linewidth]{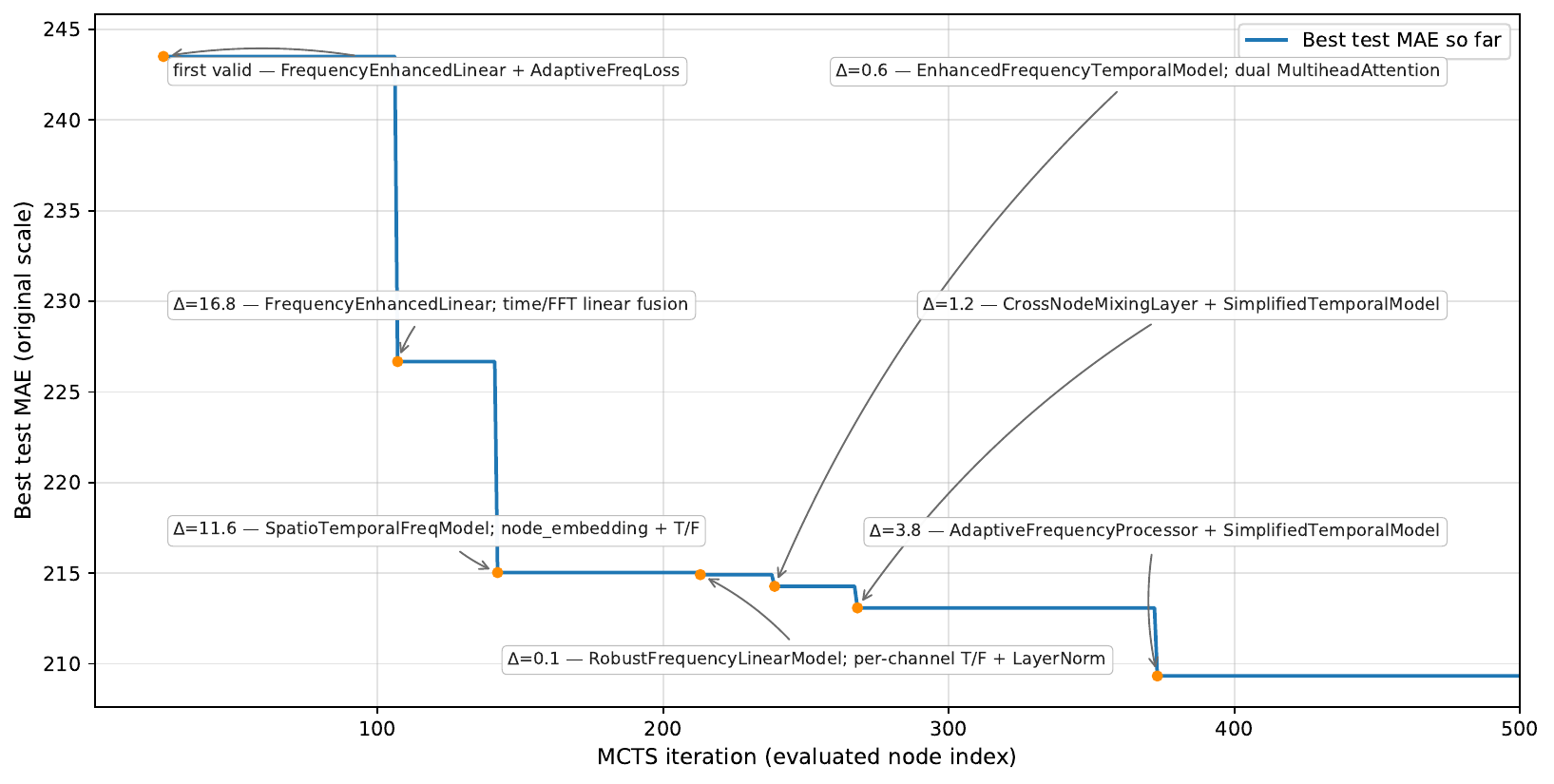}
\caption{Self evolution of SEATS generated code in load forecasting task on public ECL dataset \emph{without} reference code. Orange markers indicate iterations where the global best metric improves.}
\label{fig:ecl_mcts_progress}
\end{figure}

\paragraph{Search trajectory (ECL, w/o ref).}
\Cref{fig:ecl_mcts_progress} illustrates that progress is driven by self evolution of SEATS generated code without human intervention, most iterations leave the best metric unchanged, while a small set of iterations yields MAE reductions. The largest drop typically occurs once a first valid solution passes the metric guard, later improvements are smaller but still meaningful, and the final running best matches the ``Ours w/o ref'' ECL row in \Cref{tab:main_results}.

\begin{figure}[htbp]
\centering
\includegraphics[width=\linewidth]{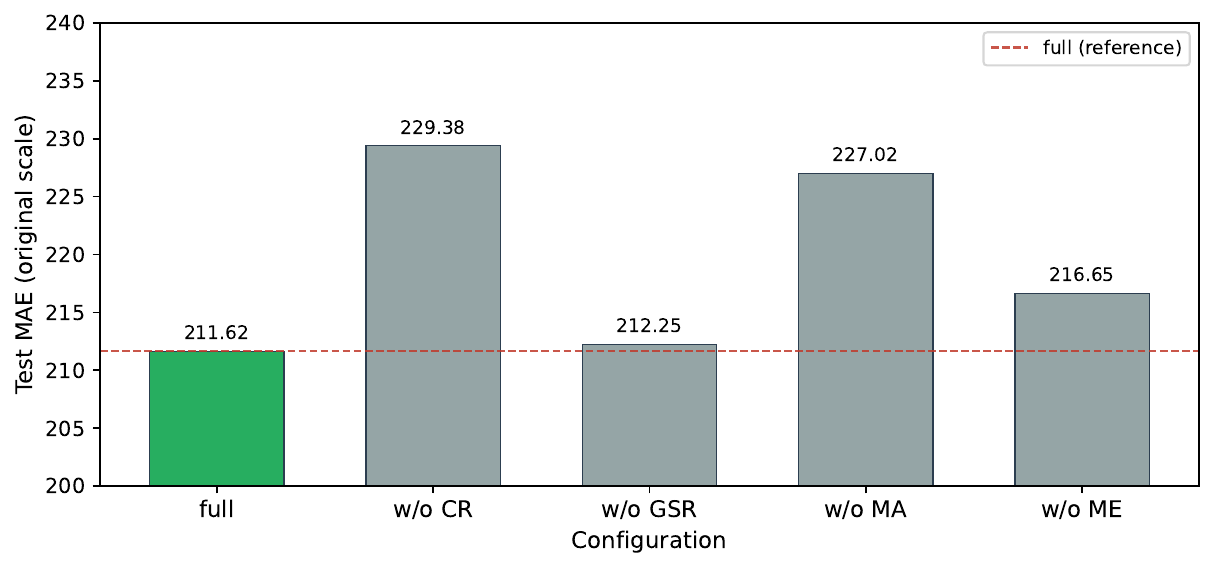}
\caption{Ablation on ECL test MAE (original scale). CR\,=\,code review, GSR\,=\,global steerable reasoning, MA\,=\,metric advantage, ME\,=\,MAP-Elites. The full system achieves the best MAE, removing CR or MA degrades performance most.}
\label{fig:ablation_ecl}
\end{figure}

\subsubsection{Ablation Study}

We ablate core \methodname{} components on the public ECL benchmark. \Cref{fig:ablation_ecl} reports test MAE when removing code review (CR), global steerable reasoning (GSR), metric advantage (MA), or MAP-Elites (ME). The full configuration attains the lowest MAE, dropping CR or MA hurts most, consistent with the role of review in suppressing leakage-style bugs and of advantage reward in focusing search.

\subsubsection{Representative SEATS Generated Models}

We briefly summarize algorithm structure generated by the search, full excerpts are listed in \Cref{app:code}.

\paragraph{Public Solar-Energy.}
The search produced \textit{DiurnalMultiScaleGatedLinear}: a six-head gated architecture with multi-scale seasonal--trend decomposition and MLP-Mixer blocks~\cite{tolstikhin2021mlpmixer}, trained with daylight-weighted Huber loss on MAD-normalized inputs (\Cref{lst:public_solar}).

\paragraph{Industry Solar PV.}
The experiment yielded \textit{QuadHeadGatedSolar}, combining residual convolutional processing, Fourier and period experts, monotonic decay for afternoon physics, horizon attention, and station--hour softmax gating with optional priors (\Cref{lst:industry_solar}).

\paragraph{ECL Load.}
The archived ECL solution uses a lightweight decomposition forecaster (\textit{SimpleDLinearModel}) with causal moving averages and MAE training on scaled load, metrics are reported after inverse scaling (\Cref{lst:ecl_load}).

\paragraph{Industry Load.}
For proprietary load, SEATS discovers a graph-attention stack over user embeddings with temporal encoding and a learnable 24-hour bias applied relative to prediction magnitude (\Cref{lst:industry_load}). An introduction of expert heads for industry solar appears in \Cref{tab:heads}.

Further qualitative discussion appears in \Cref{app:discussion}.

\section{Conclusion and Future Work}
\label{sec:conclusion}

We presented \methodname{}, a self-evolving MLE agent framework for autonomous time series forecasting algorithm development. By integrating Metric-Advantage MCTS, code review with running prompt refinement, global steerable reasoning, and quality-diversity optimization, \methodname{} autonomously generates forecasting code that is competitive with or better than strong published baselines on public and industry dataset of solar energy forecasting and load forecasting. SEATS generates diverse architectures---multi-scale gated mixers, gated solar experts with physics-motivated heads, decomposition forecasters on public load data, and graph-based residential models with learnable calibration---illustrating that autonomous code evolution can explore nontrivial design spaces beyond a single manual template.

Future work will focus on:
\begin{enumerate}
    \item \textbf{Multi-objective optimization:} Jointly optimizing accuracy, inference latency, and model size for deployment-aware generation.
    \item \textbf{Context pruning:} Reducing prompt length and API cost via tools such as SWE-Pruner~\cite{wang2026swe}.
    \item \textbf{Automated archive dimensions:} Learning MAP-Elites feature definitions from task metadata rather than hand specification.
    \item \textbf{Search algorithms:} Studying PUCT, Thompson sampling, and adaptive branching as alternatives to vanilla UCT.
\end{enumerate}

\paragraph{Broader impacts.}
Improved solar and load forecasts can support grid reliability, renewable integration, and operational planning.

\begin{ack}
\end{ack}

\bibliographystyle{unsrtnat}
\bibliography{references}

\appendix

\section{Self-Evolution Procedure}
\label{app:algorithm}

Algorithm~\ref{alg:self_evolution} formalizes the complete procedure.

\begin{algorithm}[htbp]
\caption{\fullname{} (\methodname{})}
\label{alg:self_evolution}
\begin{algorithmic}[1]
\REQUIRE Base prompt $\mathcal{P}_0$, dataset $\mathcal{D}$, budget $T$, exploration constant $C$
\STATE Initialize tree with root $N_0$ (reference code), $\mathcal{P}_{\text{run}} \leftarrow \emptyset$, $\mathcal{M} \leftarrow \emptyset$
\STATE Initialize MAP-Elites archive $\mathcal{A}$ with dimensions $(d_1, d_2, d_3)$
\FOR{$t = 1$ to $T$}
    \STATE \textbf{Phase 1 -- Node Selection:}
    \STATE Select leaf $N_{\text{parent}}$ via $\text{UCT}(N_j) = Q_j/n_j + C\sqrt{\ln n_{\text{parent}}/n_j}$
    
    \STATE \textbf{Phase 2 -- Prompt Assembly \& Code Generation:}
    \STATE $\mathcal{P}_{\text{full}} \leftarrow \mathcal{P}_0 \oplus \mathcal{P}_{\text{run}} \oplus \textsc{Context}(N_{\text{parent}}) \oplus \mathcal{P}_{\text{global}} \oplus \mathcal{P}_{\text{archive}}$
    \STATE $c_j \leftarrow \textsc{LLM}(\mathcal{P}_{\text{full}})$
    
    \STATE \textbf{Phase 3 -- Sandbox Execution \& Evaluation:}
    \STATE Execute $c_j$ in sandbox, obtain metric $M_j \leftarrow \mathcal{L}(c_j(\mathcal{D}_{\text{train}}), \mathcal{D}_{\text{test}})$
    \STATE $\mathcal{M} \leftarrow \mathcal{M} \cup \{M_j\}$
    
    \STATE \textbf{Phase 4 -- Code Review \& Prompt Update (all non-buggy nodes):}
    \IF{$N_j$ executed without errors}
        \STATE $b_j \leftarrow \textsc{CodeReview}(c_j)$ \COMMENT{Logical review}
        \STATE Compare $N_j$ with $N^*$, $N_\bot$ via Global Steerable Reasoning
        \STATE Update $\mathcal{P}_{\text{run}}$ with review findings and reasoning insights
    \ELSE
        \STATE $b_j \leftarrow \text{true}$
    \ENDIF
    
    \STATE \textbf{Phase 5 -- Tree Update:}
    \STATE Compute $A_j \leftarrow (\mu(\mathcal{M}) - M_j)/\sigma(\mathcal{M})$ \COMMENT{Metric Advantage}
    \STATE $R_j \leftarrow \begin{cases} -1 & \text{if } b_j{=}\text{true} \\ A_j & \text{otherwise} \end{cases}$
    \STATE Backpropagate: $Q_i \leftarrow Q_i + R_j,\, n_i \leftarrow n_i + 1$ for all $N_i \in \text{path}(N_j, N_0)$
    \IF{$M_j$ is new global best and $b_j{=}\text{false}$}
        \STATE $N^* \leftarrow N_j$
    \ENDIF
    \STATE Update $N_\bot$, MAP-Elites archive $\mathcal{A}$
\ENDFOR
\RETURN Best non-buggy solution $N^*$
\end{algorithmic}
\end{algorithm}

\section{Computational Complexity Details}
\label{app:complexity_details}

\paragraph{LLM Call Complexity.}
Each search iteration generates one tree node and invokes the LLM at most four times: (1) code generation, (2) code review, (3) global steerable reasoning, and (4) running prompt refinement. Calls (2)--(4) are triggered only for non-buggy nodes. With a budget of $T$ iterations, the total number of LLM calls is bounded by $4T$. In our experiments ($T{=}500$), the observed call count is at most \textit{2{,}000} per experiment.

\paragraph{Per-Call Token Estimates.}
\Cref{tab:token_estimates} summarizes the estimated input and output token ranges per call type, measured from our experimental runs. The base prompt $\mathcal{P}_0$ constitutes the largest input component (${\sim}$30--50K tokens), while the running prompt $\mathcal{P}_{\mathrm{run}}$ grows gradually across iterations.

\begin{table}[htbp]
\centering
\caption{Estimated input/output token ranges per LLM call type.}
\label{tab:token_estimates}
\small
\begin{tabular}{lcc}
\toprule
\textbf{Call Type} & \textbf{Input Tokens} & \textbf{Output Tokens} \\
\midrule
Code generation           & 30K -- 60K & 5K -- 20K \\
Code review               & 5K -- 25K  & 1K -- 5K  \\
Global steerable reasoning & 10K -- 30K & 2K -- 5K  \\
Running prompt refinement & 5K -- 15K  & 2K -- 8K  \\
\bottomrule
\end{tabular}
\end{table}

\paragraph{Wall-Clock Time.}
The wall-clock time is dominated by \textit{sandbox execution}, i.e., training and evaluating the generated ML models. The execution time per iteration varies from minutes (for lightweight models such as gradient-boosted trees) to several hours (for deep neural networks with multi-epoch training), depending on dataset size and the algorithm complexity of the generated code. With parallel sandbox execution ($P$ concurrent workers), the total wall-clock time for a full experiment is typically within \textit{one week}. Algorithmic overhead (UCT selection, backpropagation, archive updates) is negligible in comparison.

\section{Experimental Task Summary}

\begin{table}[htbp]
  \centering
  \caption{Summary of experimental tasks, datasets, and evaluation setup. Comparisons use TimeMixer~\cite{wang2024timemixer}, Timer~\cite{liu2024timer}, and SEMixer~\cite{zhang2026semixer} as external baselines.}
  \label{tab:baselines}
  \begin{tabular}{@{}llll@{}}
    \toprule
    \textbf{Task} & \textbf{Dataset} & \textbf{Primary metrics} & \textbf{Reference} \\
    \midrule
    Solar PV & Solar-Energy (Public) & MAE, MSE & \cite{lai2018modeling,wang2024timemixer} \\
    Solar PV & Proprietary PV & MAE, MSE & \cite{wang2024timemixer} \\
    Load & ECL (Public) & MAE, MSE & \cite{zhou2021informer,wang2024timemixer} \\
    Load & Proprietary Load & MAE, MSE & \cite{wang2024timemixer,liu2024timer} \\
    \bottomrule
  \end{tabular}
\end{table}

\section{Deep Research Workflow}
\label{app:deep_research}

Before each experiment, we leverage Gemini 3.1 Pro, SciMaster~\cite{chai2025scimaster}, and Claude Opus 4.6 for deep research of domain knowledge. After each experiment, we leverage Claude Opus 4.6 to reflect failure modes of experiment history, which generates a summary for further deep research. Findings of both deep research are merged into the \emph{base prompt} for future experiments. Separately, the final \emph{running prompt} produced in each experiment is also distilled into \emph{base prompt} so that recurring corrections and successful patterns persist across experiments rather than remaining tied to a single run.

\section{Expert Heads in QuadHeadGatedSolar}

\begin{table}[htbp]
\centering
\caption{Expert heads in the evolved QuadHeadGatedSolar model (industry solar).}
\label{tab:heads}
\begin{tabular}{@{}clp{7.2cm}@{}}
\toprule
\textbf{In Base Prompt} & \textbf{Expert Head} & \textbf{Description} \\
\midrule
\multirow{4}{*}{\rotatebox{90}{\scriptsize No $\star$}} 
& Monotonic Decay$^\star$ & Models afternoon solar decline via learnable amplitude, decay rate, and pivot hour with sigmoid gating, regularization enforces physical monotonicity during hours 15--18 \\
& Daily Cycle Residual$^\star$ & Per-station daily cycle profiles with fractional hour interpolation and amplitude scaling from recent history \\
& Diurnal Bias$^\star$ & Combines station embeddings, hour-of-day features, and horizon position for fine-grained bias correction \\
& Period Experts$^\star$ & Multi-period aggregation combining 24h and 12h periodic expert predictions \\
\midrule
\multirow{3}{*}{\rotatebox{90}{\scriptsize Yes}} 
& Residual TCN & Dilated depthwise temporal convolutions (kernels 3, 5, 7) \\
& Fixed-Basis Fourier & Fixed cosine/sine bases aligned to diurnal harmonics (12 frequencies) \\
& Horizon Query & Cross-attention with learnable horizon-specific queries \\
\bottomrule
\end{tabular}
\end{table}

\section{Additional Discussion}
\label{app:discussion}

\subsubsection{Autonomously Discovered Patterns}

The most compelling finding is that \methodname{} discovers novel and robust architectural patterns:

\begin{itemize}
    \item \textbf{Physics-informed constraints:} The Monotonic Decay Head encodes solar irradiance physics as a learnable architectural component with explicit regularization---a design pattern autonomously discovered without any physics-related prompting.
    \item \textbf{Robust preprocessing:} The agent consistently selected MAD-based normalization over standard scaling across multiple tasks, demonstrating emergent preference for outlier-robust methods.
    \item \textbf{Learnable hourly bias:} The magnitude-proportional bias correction in load forecasting represents a novel calibration technique.
    \item \textbf{User segmentation:} The agent independently discovered that filtering anomalous users improves aggregate prediction quality.
\end{itemize}

\subsubsection{Limitations}

\begin{itemize}
    \item Generated code quality is generally bounded by the LLM's coding capabilities. Although in base prompt we use deep research agent (such as SciMaster~\cite{chai2025scimaster}) to inject domain knowledge from papers and codebases, we notice GPT-5 (with high reasoning effort) generates more sophisticated algorithm code compared with Qwen3-coder-plus. Besides, how to conduct on-demand deep research using the signal from code generation agent more effectively to generate expert-level algorithms deserves further study.
    \item Frequent LLM calls incur high API costs. Future work will investigate context pruning~\cite{wang2026swe} to reduce token consumption.
\end{itemize}

\section{Prompt Templates}
\label{app:prompts}

This appendix provides illustrative prompt templates used in \methodname{}. The actual prompts are longer and task-specific, we show simplified versions covering all key components.

\subsection{System Prompt (Role Definition)}
\begin{lstlisting}[
    caption={System prompt defining the agent's role (simplified).},
    label=lst:system_prompt,
    language={},
    basicstyle=\ttfamily\small,
    columns=flexible,
    keepspaces=true,
    showstringspaces=false,
    frame=single,
    numbers=none,
]
You are an expert machine learning engineer.
Your task is to develop a high-performance time
series forecasting model. Generate complete,
self-contained Python code that:
1. Loads and preprocesses the provided dataset
2. Implements a novel model architecture
3. Trains the model and saves checkpoints
4. Generates predictions on the test set
5. Saves predictions to ./submission/submission.csv
Focus on designing novel and effective core model
architectures rather than writing long helper
functions.
\end{lstlisting}

\subsection{Base Prompt (Task Description)}
\begin{lstlisting}[
    caption={Base prompt template with task-specific details (simplified).},
    label=lst:base_prompt,
    language={},
    basicstyle=\ttfamily\small,
    columns=flexible,
    keepspaces=true,
    showstringspaces=false,
    frame=single,
    numbers=none,
]
## Task
Develop a forecasting model for [TASK_NAME].

## Dataset
- Format: [CSV/Parquet], Path: [DATA_PATH]
- Features: [FEATURE_LIST]
- Target: [TARGET_VARIABLE]
- Split: train [DATE1-DATE2], val [DATE3-DATE4],
         test [DATE5-DATE6]

## Evaluation Metric
[METRIC_NAME]: [FORMULA_DEFINITION]
Lower/Higher is better.

## Reference Code
[REFERENCE_CODE_EXCERPT]

## Remaining Iterations
You have [N] iterations remaining out of [T].
Consider trying underexplored approaches.

## Domain Knowledge from Deep Research 
\end{lstlisting}

\subsection{Code Review Prompt}
\begin{lstlisting}[
    caption={Code review prompt for logical error detection (simplified).},
    label=lst:review_prompt,
    language={},
    basicstyle=\ttfamily\small,
    columns=flexible,
    keepspaces=true,
    showstringspaces=false,
    frame=single,
    numbers=none,
]
## Introduction
You are an expert ML engineer reviewing code.
Identify logical errors that lead to overly
optimistic metrics, such as data leakage and
normalization issues. Focus on logic errors only.

## Generated Code
[FULL_GENERATED_CODE]

## Execution Output
[EXECUTION_LOG_AND_METRICS]

## Instructions
1. Check for data leakage: only flag leakage of
   future data from test set.
2. Check inference code correctness, especially
   checkpoint/parameter matching.
3. If you find a logical error, print the exact
   code snippet causing it.
4. For each error, review again to prove yourself
   wrong. Only flag if 100% confident.
5. Suggest prompt improvements to prevent similar
   errors in future iterations.

At the end, you MUST answer:
'has_logical_error = True' or
'has_logical_error = False'
\end{lstlisting}

\subsection{Prompt Update Prompt}
\begin{lstlisting}[
    caption={Running prompt update mechanism (simplified).},
    label=lst:update_prompt,
    language={},
    basicstyle=\ttfamily\small,
    columns=flexible,
    keepspaces=true,
    showstringspaces=false,
    frame=single,
    numbers=none,
]
## Introduction
You are a prompt engineer. Integrate the new
suggestion into the running prompt concisely.

## Current Running Prompt
[EXISTING_ACCUMULATED_INSIGHTS]

## New Suggestion for Improvement
[CODE_REVIEW_FINDINGS_OR_GLOBAL_REASONING]

## Instructions
1. Extract only actionable insights, specific
   code patterns, and concrete recommendations.
2. Organize with clear sections (e.g.,
   '## Model Architecture', '## Data Processing').
3. Prune contradictory or low-confidence info.
4. Remove redundancy with existing insights.
5. Include code snippets or pseudocode where
   relevant.
6. Output ONLY the updated running prompt.
\end{lstlisting}

\subsection{Running Prompt (Accumulated Insights)}
\begin{lstlisting}[
    caption={Running prompt example after several iterations (simplified).},
    label=lst:running_prompt,
    language={},
    basicstyle=\ttfamily\small,
    columns=flexible,
    keepspaces=true,
    showstringspaces=false,
    frame=single,
    numbers=none,
]
## Accumulated Insights
### Data Processing
- Use MAD scaling instead of standard normalization
  for robustness to sensor outliers.
- Apply .shift(1) before rolling window computations
  to prevent data leakage of future information.

### Model Architecture
- Multi-head gated architectures outperform single
  models. Consider per-station specialization.
- Physical constraints (e.g., non-negativity for
  solar, monotonic afternoon decay) improve
  generalization significantly.

### Training
- Daylight-weighted loss functions reduce error on
  practically important daytime hours.
- LR warmup + cosine annealing works well.
- Huber loss more robust than MSE for noisy data.
\end{lstlisting}

\section{Evolved Code Examples}
\label{app:code}

\noindent The listings below are excerpts from the generated solution files: 

\subsection{Public Solar-Energy: DiurnalMultiScaleGatedLinear (Excerpt)}
\begin{lstlisting}[
    caption={Multi-scale gated mixer with seasonal--trend decomposition (public Solar-Energy dataset).},
    label=lst:public_solar,
    language=Python,
    basicstyle=\ttfamily\footnotesize,
    columns=flexible,
    keepspaces=true,
    tabsize=4,
    showstringspaces=false
]
class DiurnalMultiScaleGatedLinear(nn.Module):
    def __init__(
        self,
        seq_len: int,
        pred_len: int,
        enc_in: int,
        max_freq: int = 24,
        emb_dim: int = 32,
        d_hidden: int = 128,
        dropout: float = 0.1,
        daily_period: int = 144,
    ):
        super().__init__()
        self.seq_len = seq_len
        self.pred_len = pred_len
        self.enc_in = enc_in
        self.daily_period = daily_period
        self.K = 6

        assert seq_len % 4 == 0, "seq_len should be divisible by 4"

        self.ma0 = MovingAvg(kernel_size=25)
        self.ma48 = MovingAvg(kernel_size=13)
        self.ma24 = MovingAvg(kernel_size=7)

        self.core_mixer = MixerBlock(
            seq_len, enc_in, token_mlp_dim=128, channel_mlp_dim=512, dropout=dropout
        )
        self.ds48_mixer = MixerBlock(
            seq_len // 2, enc_in, token_mlp_dim=128, channel_mlp_dim=512, dropout=dropout,
        )
        self.ds24_mixer = MixerBlock(
            seq_len // 4, enc_in, token_mlp_dim=128, channel_mlp_dim=512, dropout=dropout,
        )
        # ... projection heads, Fourier/diurnal experts, station-hour gating ...
\end{lstlisting}

\subsection{Industry Solar: QuadHeadGatedSolar (Excerpt)}
\begin{lstlisting}[
    caption={Six-head gated solar forecaster with station--hour softmax gating (industry PV).},
    label=lst:industry_solar,
    language=Python,
    basicstyle=\ttfamily\footnotesize,
    columns=flexible,
    keepspaces=true,
    tabsize=4,
    showstringspaces=false
]
class QuadHeadGatedSolar(nn.Module):
    def forward(self, x_norm, hours_future, hpos, station_ids, y_scaler, y_hist_raw, hours_hist):
        res_den = self.residual_head(x_norm, y_scaler)
        fou_den = self.fourier(y_hist_raw)
        cyc_den = self.cycle_residual_head(station_ids, hours_hist, hours_future, y_hist_raw)
        per_den = self.period_head(y_hist_raw)
        dec_den = self.decay_head(y_hist_raw, hours_future)
        hq_den = self.hq_head(x_norm, y_scaler)
        heads = torch.stack([res_den, fou_den, cyc_den, per_den, dec_den, hq_den], dim=-1)
        hrsf = hours_future.float()
        hour_feat = torch.stack([
            torch.sin(2 * math.pi * hrsf / 24.0),
            torch.cos(2 * math.pi * hrsf / 24.0),
            hpos,
        ], dim=-1)
        hour_emb = self.hour_feat_proj(hour_feat)
        st_emb = self.station_emb(station_ids).unsqueeze(1).expand(-1, hours_future.size(1), -1)
        gate_logits = self.gate_mlp(hour_emb + st_emb)
        if self.use_hour_priors:
            gate_logits = gate_logits + self.hour_prior[(hours_future % 24).long()]
        weights = torch.softmax(gate_logits / max(1e-4, self.gate_tau), dim=-1)
        y_den = torch.sum(heads * weights, dim=-1).clamp_min(0.0)
        return y_den, weights
\end{lstlisting}

\subsection{Public ECL: SimpleDLinearModel (Excerpt)}
\begin{lstlisting}[
    caption={Decomposition + linear forecast head on the public ECL benchmark.},
    label=lst:ecl_load,
    language=Python,
    basicstyle=\ttfamily\footnotesize,
    columns=flexible,
    keepspaces=true,
    tabsize=4,
    showstringspaces=false
]
class SimpleDLinearModel(nn.Module):
    def __init__(self, seq_len, pred_len, num_features, moving_avg_kernel=25):
        super().__init__()
        self.seq_len = seq_len
        self.pred_len = pred_len
        self.num_features = num_features
        self.decomposition = SeriesDecomposition(moving_avg_kernel)
        self.seasonal_pred = nn.Linear(seq_len, pred_len)
        self.trend_pred = nn.Linear(seq_len, pred_len)

    def forward(self, x):
        seasonal, trend = self.decomposition(x)
        seasonal_transposed = seasonal.transpose(1, 2)
        trend_transposed = trend.transpose(1, 2)
        seasonal_pred = self.seasonal_pred(seasonal_transposed)
        trend_pred = self.trend_pred(trend_transposed)
        seasonal_pred = seasonal_pred.transpose(1, 2)
        trend_pred = trend_pred.transpose(1, 2)
        return seasonal_pred + trend_pred
\end{lstlisting}

\subsection{Industry Load: GraphNeuralNetwork (Excerpt)}
\begin{lstlisting}[
    caption={User-graph attention with temporal encoder and hourly bias (industry load).},
    label=lst:industry_load,
    language=Python,
    basicstyle=\ttfamily\footnotesize,
    columns=flexible,
    keepspaces=true,
    tabsize=4,
    showstringspaces=false
]
class GraphNeuralNetwork(nn.Module):
    def __init__(self, input_dim, num_users, embed_dim=48, hidden_dim=128,
                 graph_layers=2, dropout=0.1):
        super(GraphNeuralNetwork, self).__init__()

        self.user_embedding = nn.Embedding(num_users, embed_dim)

        self.temporal_encoder = TemporalEncoder(
            input_dim, hidden_dim // 2, dropout=dropout
        )

        self.graph_layers = nn.ModuleList()
        gnn_input_dim = embed_dim
        for i in range(graph_layers):
            if i == graph_layers - 1:
                gnn_output_dim = hidden_dim // 2
            else:
                gnn_output_dim = embed_dim
            self.graph_layers.append(
                GraphAttentionLayer(gnn_input_dim, gnn_output_dim, dropout=dropout)
            )
            gnn_input_dim = gnn_output_dim

        self.combined_fc = nn.Sequential(
            nn.Linear(hidden_dim, hidden_dim),
            nn.LayerNorm(hidden_dim),
            nn.GELU(),
            nn.Dropout(dropout),
            nn.Linear(hidden_dim, hidden_dim // 2),
            nn.LayerNorm(hidden_dim // 2),
            nn.GELU(),
            nn.Dropout(dropout),
        )

        self.output = nn.Linear(hidden_dim // 2, 1)

        self.hour_bias = nn.Parameter(torch.zeros(24))
        with torch.no_grad():
            self.hour_bias[2:8] = 0.05
            self.hour_bias[8:21] = -0.03

    def forward(self, x_temporal, user_ids, hours):
        temporal_out = self.temporal_encoder(x_temporal)

        unique_user_ids, inverse_indices = torch.unique(
            user_ids, return_inverse=True
        )
        num_nodes = unique_user_ids.size(0)

        unique_embs = self.user_embedding(unique_user_ids)
        # ... graph attention over users, readout, combined_fc ...
        # final: out + hour_bias[hours] * |out| * scale
\end{lstlisting}
\end{document}